# Learning the Language of Histopathology Images reveals Prognostic Subgroups in Invasive Lung Adenocarcinoma Patients

Abdul Rehman Akbar[1,*], Usama Sajjad[1], Ziyu Su[1], Wencheng Li[2], Fei Xing[3], Jimmy Ruiz[4,5], Wei Chen[1], Muhammad Khalid Khan Niazi[1]

[1] Department of Pathology, College of Medicine, The Ohio State University Wexner Medical Center, Columbus, OH, USA

[2] Department of Pathology, Wake Forest University School of Medicine, Winston-Salem, NC, USA

[3] Department of Cancer Biology, Wake Forest University School of Medicine, Winston-Salem, NC, USA

[4] Department of Medicine (Hematology & Oncology), Wake Forest University School of Medicine, Winston-Salem, NC, USA

[5] Section of Hematology & Oncology, W.G. (Bill) Hefner Veterans Affair Medial Center (VAMC), Salisbury, NC, USA

[*] Corresponding Author (Abdul.Akbar@osumc.edu)

## Abstract

Recurrence remains a major clinical challenge in surgically resected invasive lung adenocarcinoma, where existing grading and staging systems fail to capture the cellular complexity that underlies tumor aggressiveness. We present **PathRosetta**, a novel AI model that conceptualizes histopathology as a language, where cells serve as words, spatial neighborhoods form syntactic structures, and tissue architecture composes sentences. By learning this language of histopathology, PathRosetta predicts five-year recurrence directly from hematoxylin-and-eosin (H&E) slides, treating them as documents representing the state of the disease. In a multi-cohort dataset of **289 patients** (**600 slides**), PathRosetta achieved an area under the curve (AUC) of **0.78±0.04** on the internal cohort, significantly outperforming IASLC grading (AUC:0.71), AJCC staging (AUC:0.64), and other state-of-the-art AI models (AUC:0.62–0.67). It yielded a **hazard ratio of 9.54** and a **concordance index of 0.70**, generalized robustly to external TCGA (AUC:0.75) and CPTAC (AUC:0.76) cohorts, and performed consistently across demographic and clinical subgroups. Beyond whole-slide prediction, PathRosetta uncovered **prognostic subgroups within individual cell types**, revealing that even within benign epithelial, stromal, or other cells, distinct morpho-spatial phenotypes correspond to divergent outcomes. Moreover, because the model explicitly understands what it is looking at, including cell types, cellular neighborhoods, and higher-order tissue morphology, it is inherently interpretable and can articulate the rationale behind its predictions. These findings establish that representing histopathology as a language enables interpretable and generalizable prognostication from routine histology.

## Introduction

Lung adenocarcinoma, the most common histological subtype of non-small cell lung cancer (NSCLC), remains a leading cause of cancer-related death worldwide, accounting for over 40% of NSCLC cases and contributing to more than 1.7 million deaths annually [1, 2]. Among these, the majority (~70%) are invasive lung adenocarcinoma (ILA), which represent the focus of this study. Despite advances in early detection and surgical management, patient outcomes remain poor. Five-year survival rates range from 70–90% for stage I disease to below 10% for stage IV, reflecting the strong stage dependence of prognosis [3, 4]. However, even among patients with early-stage disease treated with curative-intent resection, 20–30% experience recurrence within five years, and this rate exceeds 50% in stage III [5]. These statistics underscore the urgent need for reliable prognostic tools to identify high-risk patients and guide postoperative management [6].

However, current prognostic tools offer alarmingly limited predictive value. The tumor node metastasis (TNM) staging system, considered the gold standard for treatment decisions, achieves only modest prognostic performance for overall survival prediction [7-9]. Even more concerning, studies specifically evaluating recurrence prediction show that TNM staging achieves an area under the curve (AUC) of merely 0.561, while tumor grading performs only marginally better at 0.573 [10]. Histologic subtyping, despite recent advances in the International Association for the Study of Lung Cancer (IASLC) grading system, demonstrates AUCs ranging from 0.68-0.70 for recurrence-free survival [11]. These disappointing results stem from inherent limitations including interobserver variability and inability of these coarse-grained systems to capture the complex biological heterogeneity of tumors [12, 13]. There is thus an urgent need for more robust, biologically informed histopathological signals that can capture the complex cellular and architectural features underlying tumor heterogeneity, which when integrated with molecular and clinical data, could enable the development of comprehensive prognostic tools with superior predictive accuracy.

The advent of digital pathology has revolutionized cancer research by enabling computational analysis of whole slide images (WSIs) of hematoxylin and eosin (H&E)-stained tissue sections, which contain rich information about tumor architecture, cellular morphology, and the tumor microenvironment (TME) that may harbor high-dimensional prognostic signals invisible to the naked eye [14, 15]. However, extracting meaningful signals from these gigapixel-sized images remains a significant challenge.

The application of artificial intelligence (AI) in pathology has shown significant potential, not only for diagnosing tumors but also for predicting diverse clinical endpoints [15-24]. However, the majority of these AI approaches rely on foundation models (FMs), task-agnostic architectures trained on massive datasets in a self-supervised manner [25-29]. Natural images, the primary domain for which these paradigms were developed, lack inherent hierarchical structure. A photograph of a landscape or street scene contains pixels and regions, but these do not naturally organize into semantically defined building blocks. Modern vision transformers encode such images by dividing them into patches and learning relationships through self-attention, treating each patch as a token. Self-supervised methods such as DINO (Self-Distillation with No labels) [30] further refine this approach by encouraging similarity between different views of the same image. While these paradigms have demonstrated impressive success in general computer vision, their direct transfer to histopathology remains challenging. These limitations hinder FM performance on prognostic tasks such as survival and recurrence risk prediction, particularly when used as frozen feature extractors (Fig. 1A) [31].

In stark contrast, natural language processing has been revolutionized by an architectural insight: language possesses inherent hierarchical building blocks. Alphabets combine to form words, words assemble into phrases, phrases construct sentences, and so forth. Language models explicitly encode this structure, capturing meaning at multiple scales simultaneously (Fig. 1B). This hierarchical encoding, where the meaning of each element is determined by its context and position within the broader structure, has enabled unprecedented success in specialized domains.

Remarkably, histopathology represents a profound exception among visual domains: it possesses intrinsic biological building blocks analogous to linguistic structures. Individual cells serve as fundamental "words," cellular neighborhoods function as "phrases," tissue regions act as "sentences," and whole slides constitute complete "documents." Just as the meaning of a word shifts based on context, the biological significance of a cell is determined by its spatial neighbors and microenvironment. Despite this compelling analogy, computational pathology remains largely anchored to vision models designed for natural images, fundamentally overlooking the structured, hierarchical nature of biological tissues.

To fill this gap, we introduce PathRosetta, an AI framework that models the “language of histopathology” by explicitly treating tissue as a structured biological text. To the best of our knowledge, this represents the first study to explicitly leverage hierarchical, language-inspired modeling principles for computational pathology. In PathRosetta, individual cells act as fundamental tokens, their interactions and spatial neighborhoods form phrases, and higher-order tissue regions assemble into sentences, collectively defining the grammar of disease (Fig. 1C). By learning how cells “communicate” through their spatial and morphological context, PathRosetta captures the cellular syntax and semantic relationships that govern tumor architecture and microenvironmental organization. This biologically grounded representation enables robust outcome prediction and offers interpretable insights into the cellular and structural dynamics that shape disease behavior.

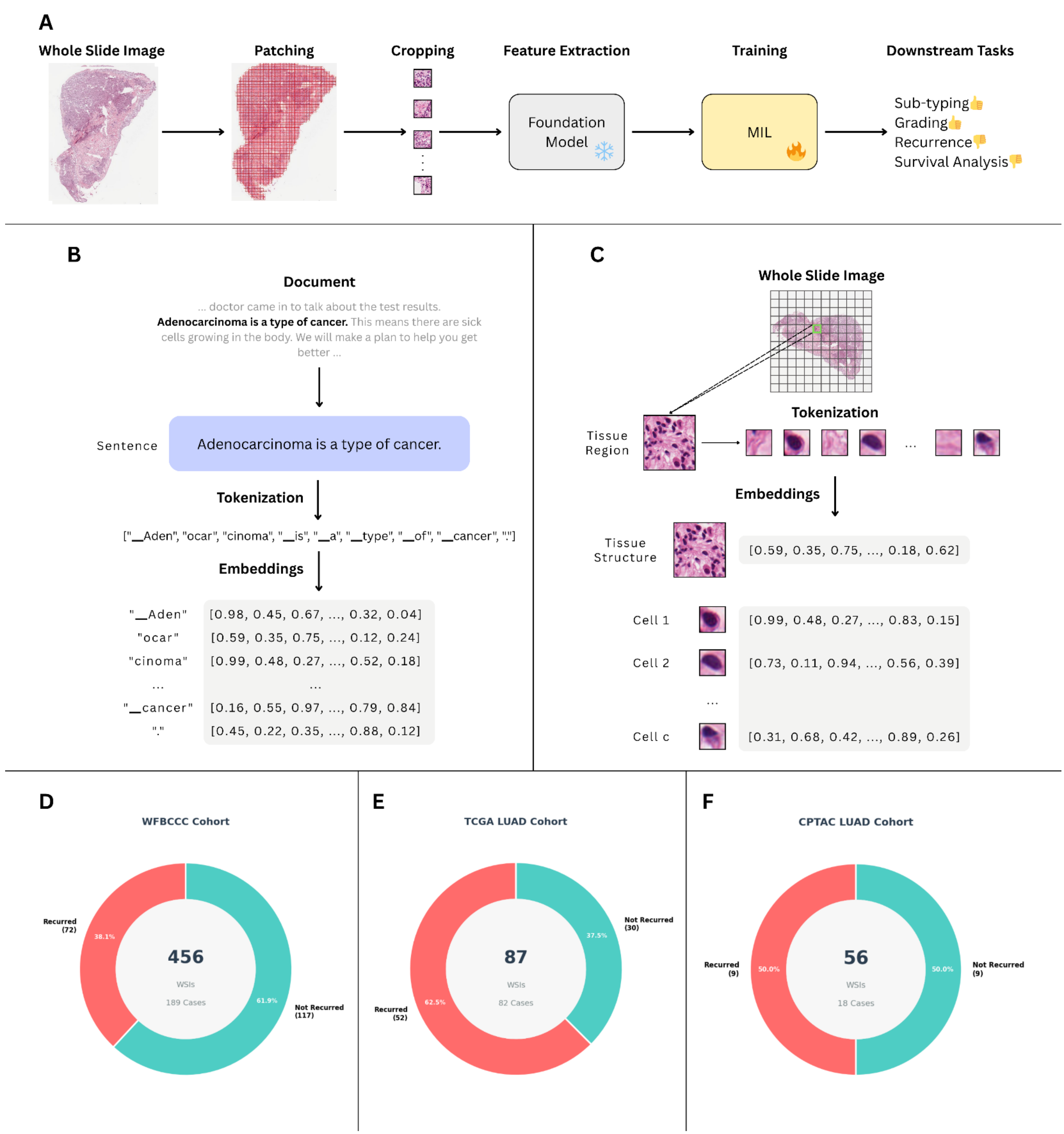

**Figure 1 | Conceptual overview of PathRosetta and study cohorts (A)** Traditional computational pathology workflows treat a WSI as a set of patches that are cropped, processed through a foundation model, and aggregated using MIL for downstream tasks such as subtyping, grading, recurrence prediction, or survival

analysis. **(B)** In natural language processing, meaning emerges from word arrangement and context. Language models encode this structure by tokenizing text and representing words as contextual embeddings that capture semantics and syntax. **(C)** PathRosetta applies this linguistic principle to histopathology by tokenizing tissue into individual cells, learning their morphological and spatial embeddings, and modeling cell–cell interactions as biological "sentences" that encode tumor behavior. **(D-F)** Cohorts used for training and validation: (D) internal WFBCCC cohort (456 WSIs, 189 cases, 38.1% recurrence); (E) TCGA-LUAD external test set (87 WSIs, 82 cases, 42.7% recurrence); and (F) CPTAC-LUAD external test set (56 WSIs, 18 cases, 50% recurrence). Together, these datasets enabled development and cross-institutional validation of PathRosetta as a framework that models the language of pathology.

## Results

### PathRosetta outperforms clinical and computational benchmarks

We evaluated PathRosetta for predicting 5-year recurrence in a cohort of 189 patients with ILA, using 456 H&E-stained WSIs (Fig 1D). Under a rigorous five-fold patient-wise cross-validation framework, PathRosetta substantially outperformed both established clinicopathologic variables and state-of-the-art AI models. Traditional baselines, including AJCC stage, dominant pattern–based grading, and the IASLC grading system, demonstrated limited prognostic power, with AUCs ranging from 0.64 to 0.71 and concordance indices of 0.63–0.67 (Fig. 2A). Similarly, leading AI models achieved a maximum AUC of just 0.67. In contrast, PathRosetta achieved an AUC of 0.78, representing a greater than 16% relative improvement over second-best AI method (AUC: 0.67) along with an accuracy of 71.7%, a sensitivity of 65.8%, and a specificity of 75.5% (Fig. 2B, Supplementary Table S2).

A

Methods
AJCC Stage
Prominent Pattern Based Grading System
IASLC Grading System
PathRosetta

Performance Score
0.8
0.75
0.7
0.65
0.6

0.6401
0.6485
0.7144
0.7784
Area Under the Curve (AUC)

0.6064
0.6026
0.6655
0.7171
Accuracy

0.6363
0.6309
0.6726
0.7080
Concordance Index (C-index)

Performance Metrics

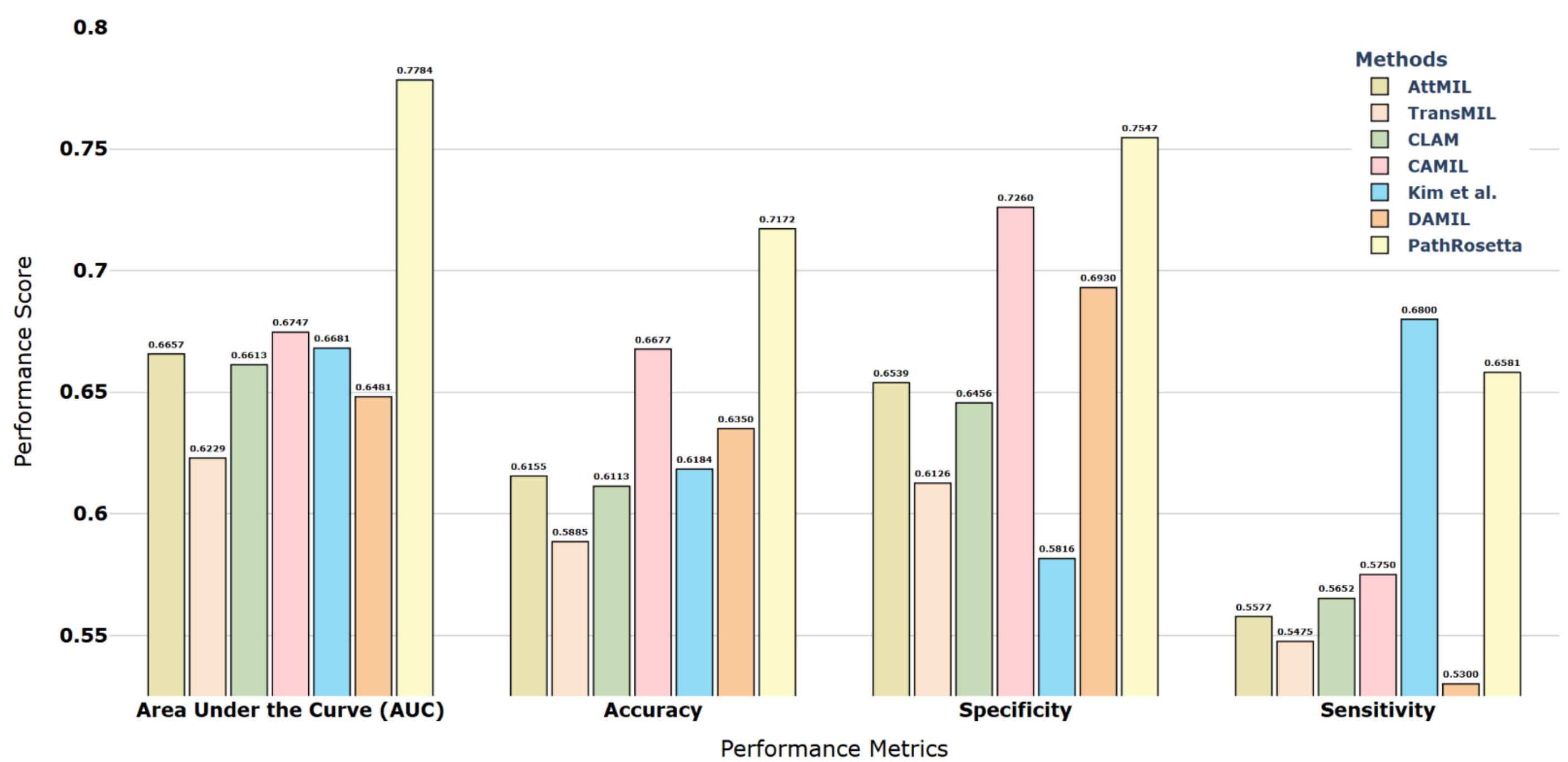

**Figure 2 | PathRosetta comparison with clinical and computational benchmarks (A)** Comparison of PathRosetta with established clinicopathologic prognostic systems, including AJCC stage, the prominent pattern–based grading system, and the IASLC grading system. **(B)** Comparison with leading AI models, including AttMIL, TransMIL, CLAM, CAMIL, DAMIL, and Kim et al.

**Cell-type–specific language modeling reveals prognostically distinct subgroups within pathologically defined cell types**

PathRosetta's flexible architecture enables the construction of tissue "documents" from different cellular vocabularies. By treating each cell type as a distinct lexicon, we can generate separate histopathology languages, one composed solely of stromal cell "words," another from inflammatory cells, and so forth. To investigate which cellular vocabularies carry the most prognostic information for recurrence prediction, we trained separate models using five cell-type–specific languages: stromal, inflammatory, neoplastic, benign epithelial, and dead cells. Each demonstrated distinct yet complementary prognostic value revealing subgroups within a cell type with divergent outcomes. Among these, the benign epithelial cell model performed best (AUC = 0.75), followed by dead cells (AUC = 0.73) and stromal cells (AUC = 0.72), while inflammatory and neoplastic cell models also showed solid discriminative ability (AUCs = 0.70–0.72). An All-Cell model, trained on pooled cellular features without distinguishing types, achieved an AUC of 0.74, while the combined ensemble model, integrating predictions across all cell-type–specific models, achieved the highest performance (AUC = 0.78, accuracy = 71.7%, specificity = 75.5%) (Fig. 3, Supplementary Table S3).

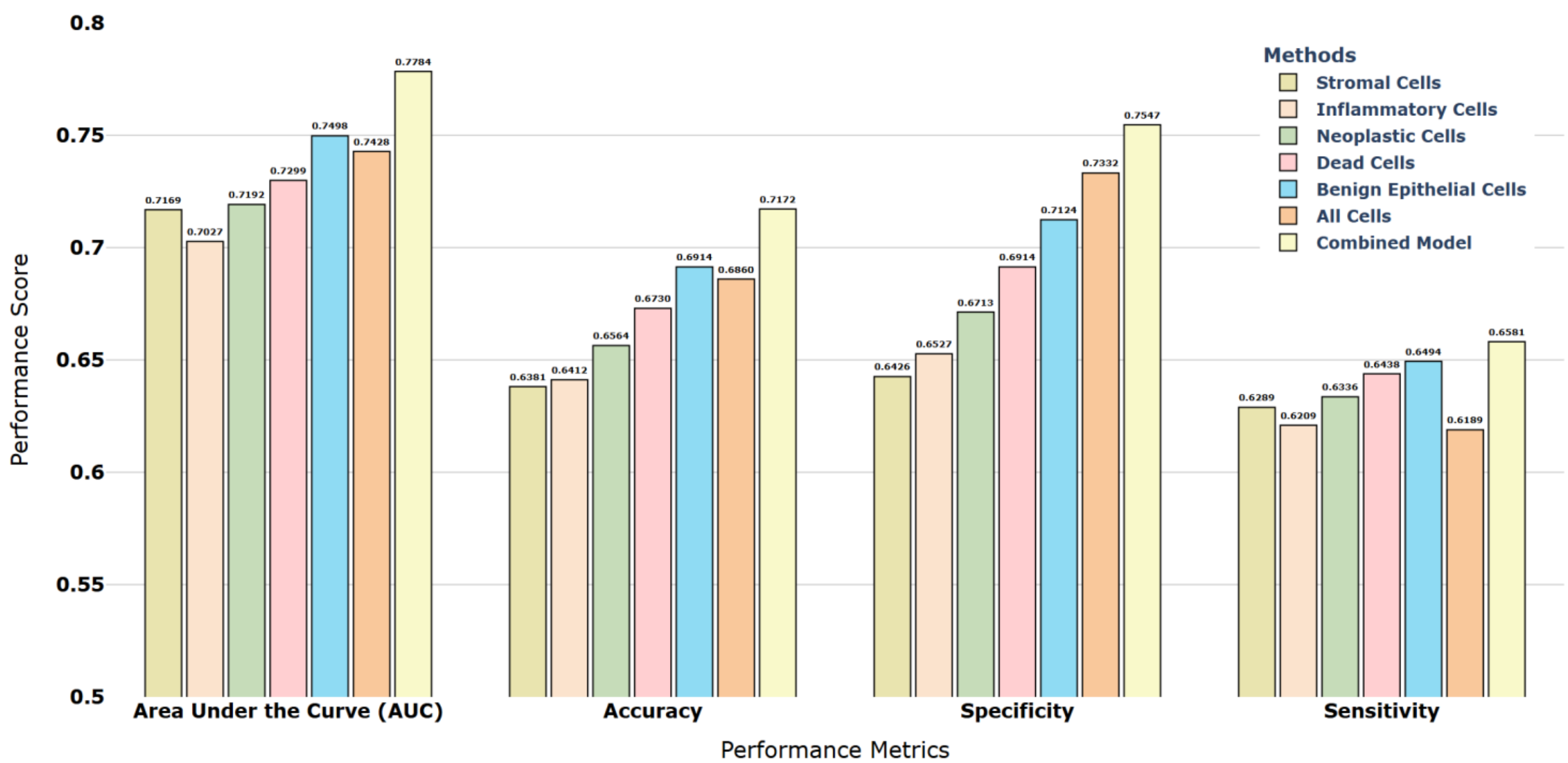

**Figure 3 | Cell-type–specific and ensemble model performance** Performance comparison of PathRosetta models trained on individual cell types, the all-cell model, and the combined ensemble model for 5-year recurrence prediction in ILA. Each bar represents the mean performance across five cross-validation folds for four key metrics: area under the curve (AUC), accuracy, specificity, and sensitivity.

**Risk stratification and survival analysis**

We next assessed clinical utility using Cox proportional hazards regression. PathRosetta achieved the strongest risk stratification, with a hazard ratio (HR) of 9.54 (95% CI: 4.34–20.98, $p < 0.005$), indicating a more than ninefold higher recurrence risk for patients predicted as high-risk compared to low-risk. The All-Cell model also demonstrated substantial prognostic separation (HR = 7.70, 95% CI: 3.80–15.60, $p < 0.005$) (Supplementary Table S4). Kaplan–Meier analyses confirmed clear and statistically significant separations between high- and low-risk groups across all models ($p < 0.0001$; Fig. 4). The benign epithelial and dead cell models achieved the strongest stratification among individual cell types (HRs = 6.04 and 5.69, respectively), consistent with their superior AUC and concordance index values (Supplementary Table S4).

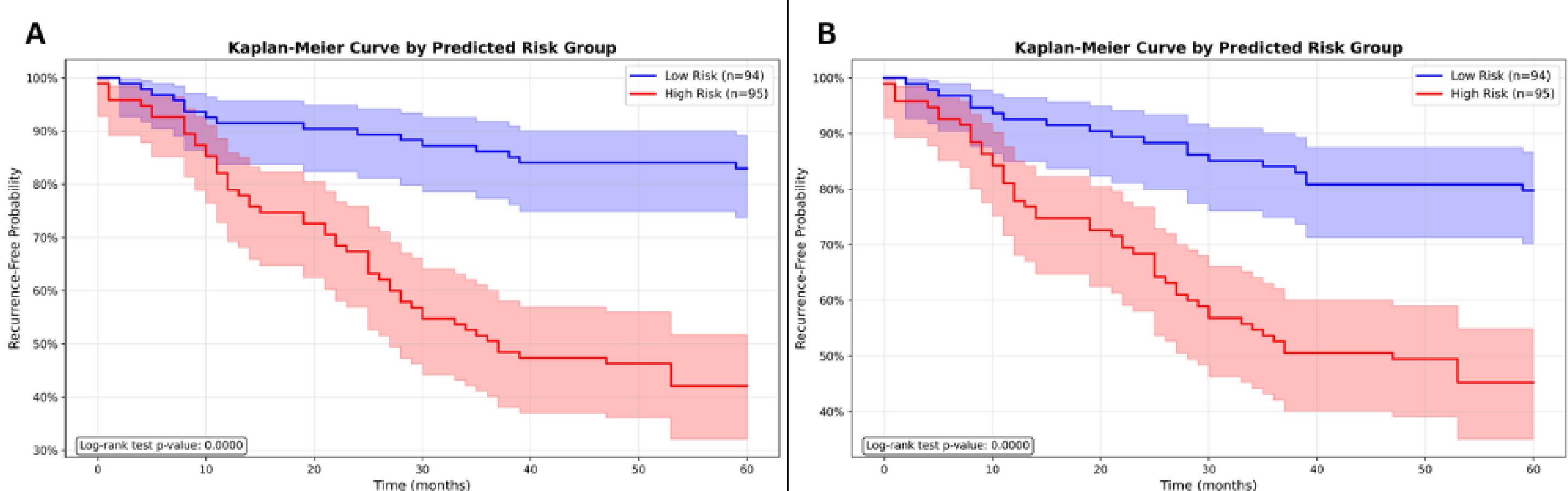

**Figure 4 | Kaplan–Meier survival analysis based on predicted recurrence risk** Kaplan–Meier recurrence-free survival curves for patients stratified into high- and low-risk groups based on model predictions. **(A)** The All-Cells model trained using pooled cellular features without distinguishing between cell types. **(B)** The Combined ensemble model, which integrates predictions from all cell-type–specific models.

**PathRosetta demonstrates consistent performance across demographic and clinical subgroups**

To ensure fairness and robustness, we analyzed PathRosetta's performance across demographic and clinical subgroups, including sex, race, age, tumor stage, and histologic grade. No significant differences in false negative rates were observed across sex (female = 11.2%, male = 15.6%; p = 0.52), race (White = 12.8%, African American = 10.0%; p = 0.48), or age groups (≥65 years = 14.1%, <65 years = 12.4%; p = 0.84). There was also no statistically significant difference between recurrence and death timings of the patients (Fig. 5). Performance trends followed clinical expectations, with higher AUCs in well-differentiated (IASLC Grade 1, AUC = 0.87) and early-stage (Stage I, AUC = 0.75) tumors, and lower scores in advanced disease (Stage III, AUC = 0.53). The diminished accuracy in Stage III cases likely reflects small sample size (n = 15) and uneven fold representation (Supplementary Table S5). These analyses were done on the respective test sets of each fold's trained model.

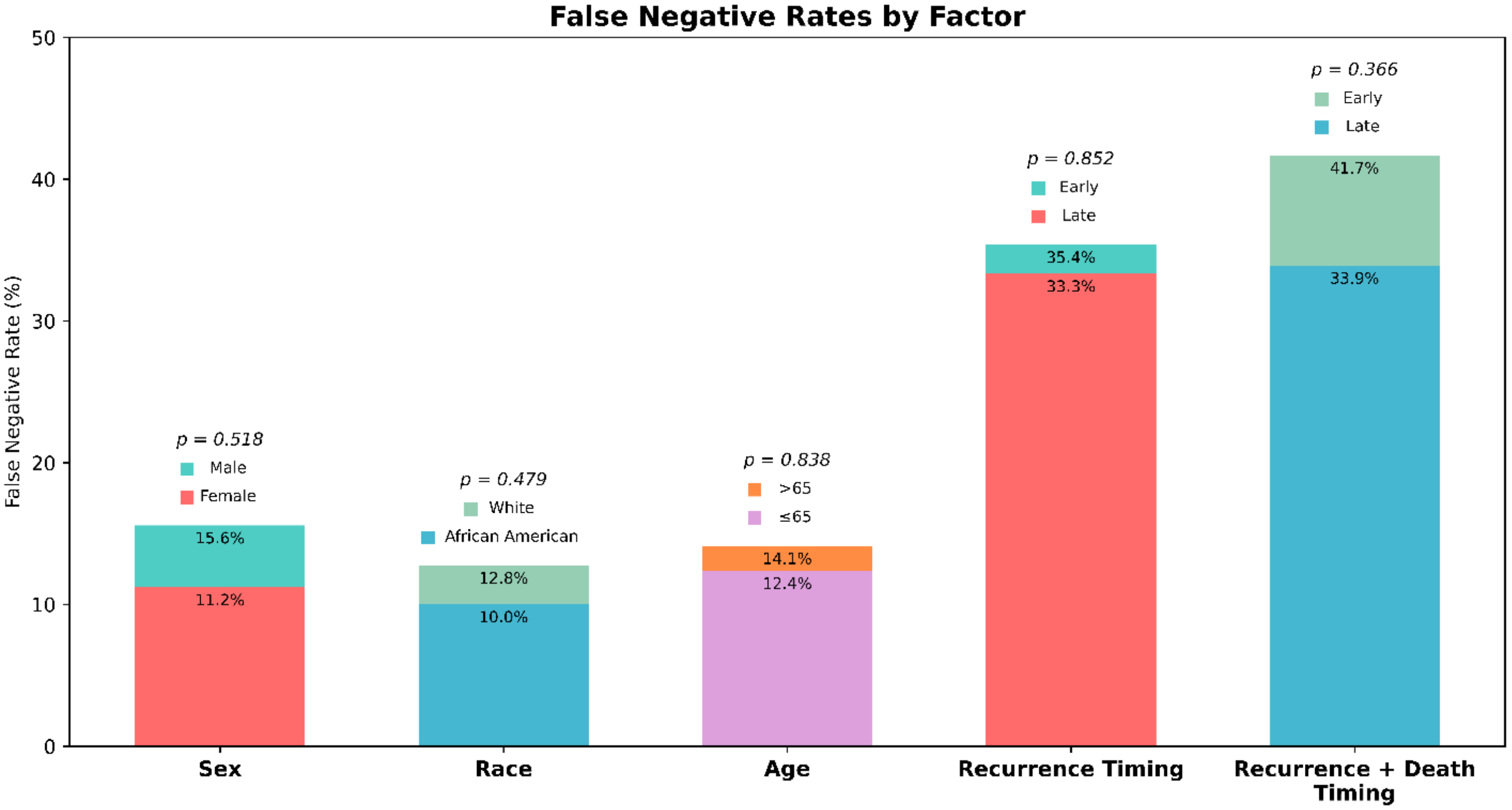

**Figure 5 | Evaluation of model fairness and failure patterns across demographic and clinical subgroups**

False negative rates for the PathRosetta were compared across demographic (sex, race, age) and clinical (recurrence timing, recurrence + death timing) subgroups. Bars represent mean false negative rates within each group, and p-values indicate statistical test results for group differences.

**External validation confirms generalizability**

To evaluate the generalizability of PathRosetta beyond the development cohort, we tested the model on two independent external datasets: the TCGA-LUAD (Fig. 1E) and CPTAC-LUAD (Fig. 1F) cohorts. Despite variations in slide preparation, staining, and scanning protocols, PathRosetta maintained strong predictive performance, achieving AUCs of 0.75 and 0.76 on TCGA and CPTAC, respectively. The model also demonstrated balanced accuracy and specificity exceeding 72% across both datasets, indicating robust discrimination between patients with and without recurrence (Fig. 6). Detailed quantitative comparisons with other methods are provided in Supplementary Table S6.

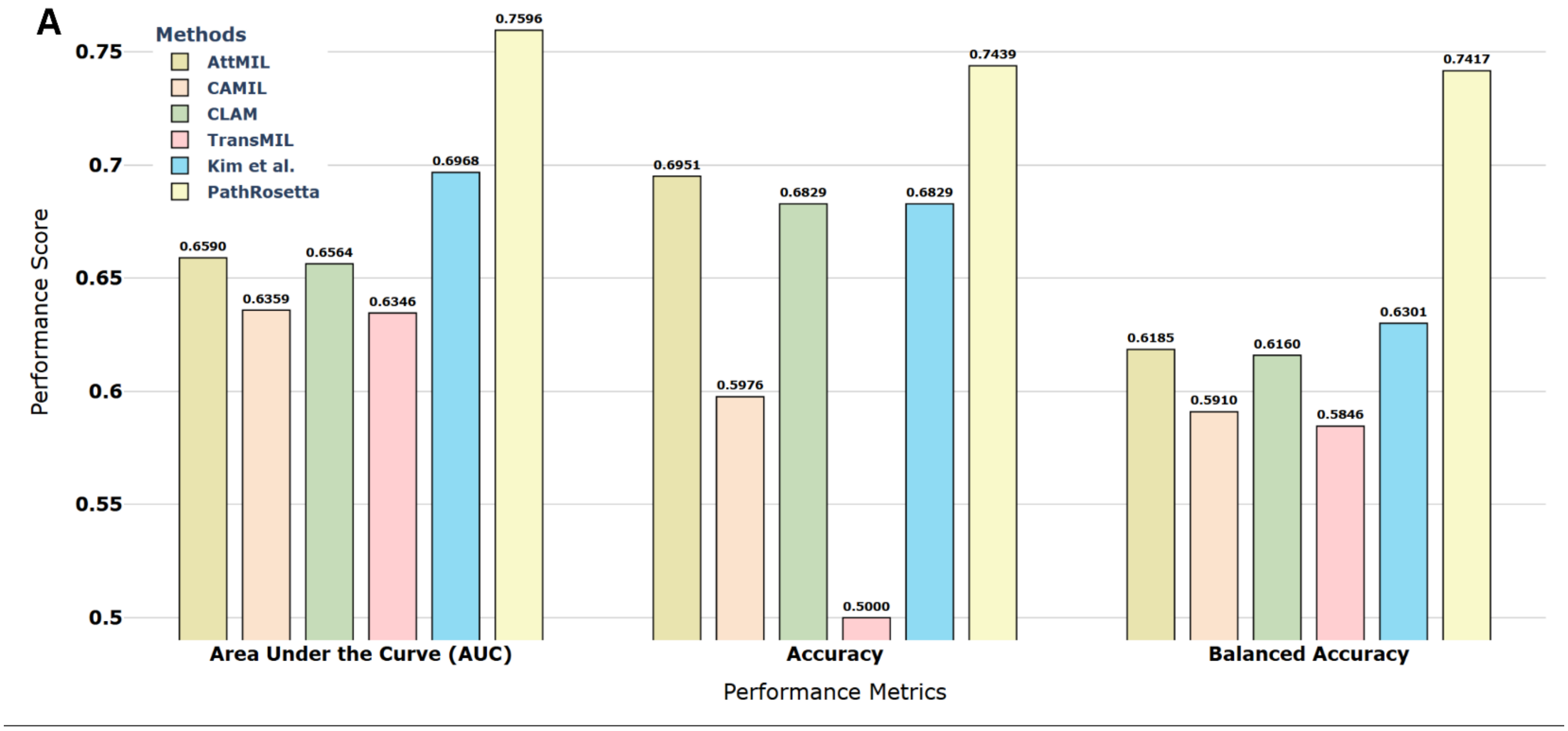

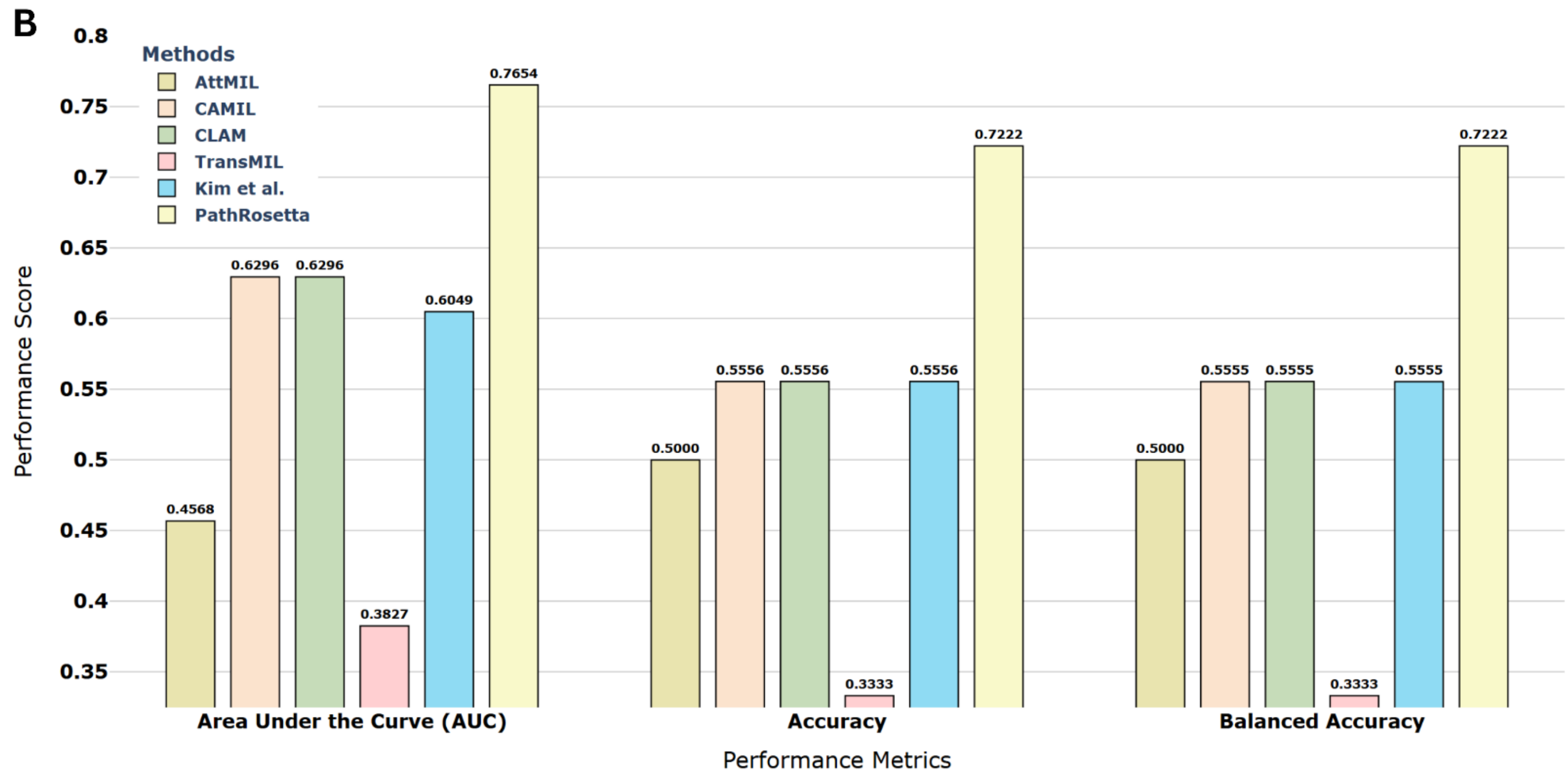

**Figure 6 | External validation of PathRosetta on independent LUAD cohorts** Performance comparison of PathRosetta with leading AI models on two independent datasets. **(A)** Results on the TCGA-LUAD cohort (82 cases, 87 WSIs) showing that PathRosetta achieves the highest performance across all metrics, outperforming all comparative methods, including AttMIL, CAMIL, CLAM, TransMIL, and Kim et al. **(B)** Results on the CPTAC-LUAD cohort (18 cases, 56 WSIs), where PathRosetta again demonstrates superior generalization despite variations in slide preparation and scanning protocols.

## Discussion

Traditional histopathology models treat WSIs as collections of patches, overlooking the basic biological reality that cancer is a cellular disease [32]. Tumors are composed of heterogeneous cell populations that differ subtly in morphology, molecular state, and function. Prior studies have shown that features such as nuclear shape, texture, and size, often imperceptible to the human eye, carry strong prognostic value [33, 34]. Moreover, TME includes dynamic networks of stromal, immune, and epithelial cells that modulate disease progression and treatment response. Together, these findings highlight that accurate outcome prediction requires modeling not only the presence of specific cells, but also their spatial context and interactions.

PathRosetta operationalizes this concept by treating individual cells as the basic “tokens” of a histopathological language. Analogous to how language models derive meaning from the arrangement of words, PathRosetta learns how spatial and morphological relationships among cells form the biological “sentences” that encode the state of the tumor ecosystem. The model captures not only what cells exist, but the context in which they coexist, information that patch-based methods inherently dilute.

To illustrate this principle, we analyzed cell-level embeddings extracted using CellViT++ [35, 36] for five major cell categories. Visualization of cell embeddings revealed subgroups within benign epithelial cells of patients with and without recurrence, even though the cells appeared histologically similar (Fig. 7, Supplementary Fig. S1). This visual separation within the subgroups of same cell type reflects hidden molecular or functional states that influence disease behavior, emphasizing that cancer is not merely a disorder of morphology but a complex system of cellular states and communications. By operating in high-dimensional feature spaces, PathRosetta captures this latent heterogeneity, detecting subtle, prognostically relevant differences invisible to naked eye.

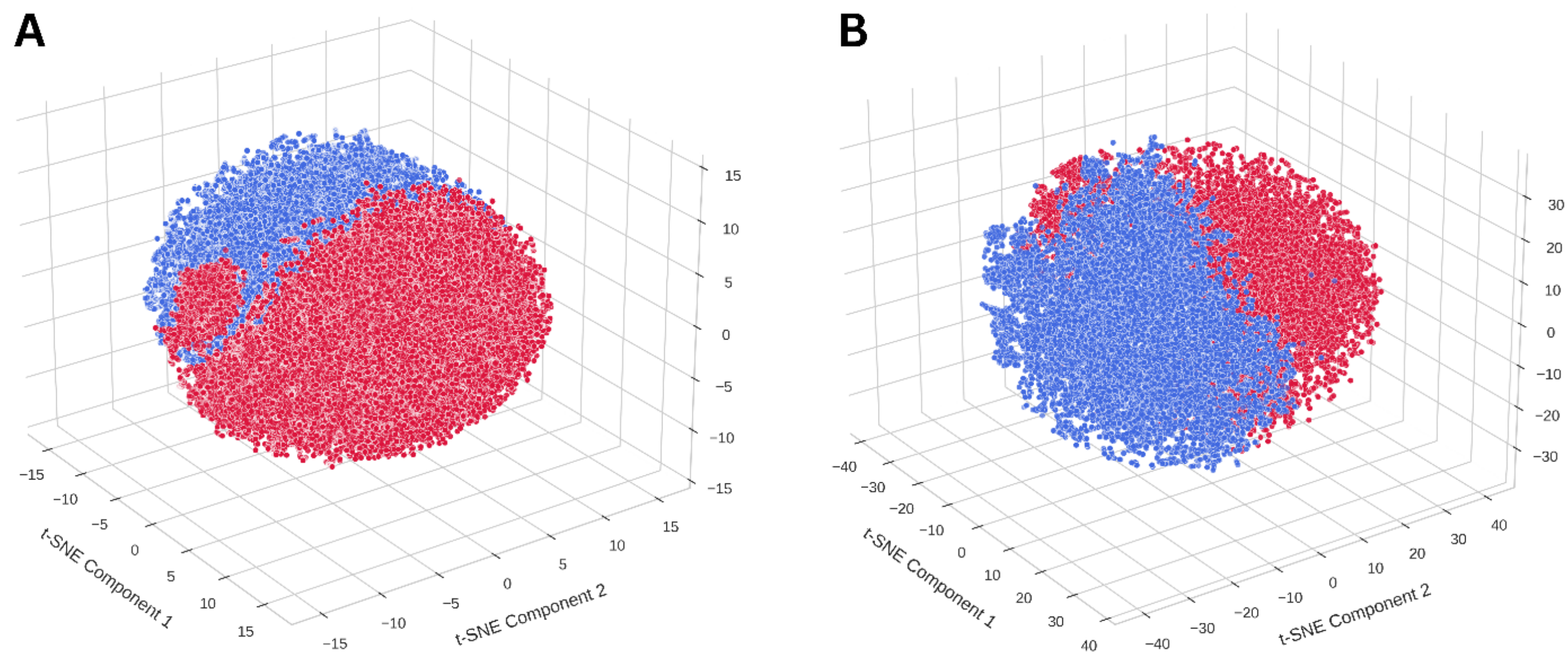

**Figure 7 | Visualization of sub-visual cellular heterogeneity in benign epithelial cells** Three-dimensional t-SNE projections of cell-level embeddings extracted using CellViT++ for benign epithelial cells from recurrence-positive (red) and recurrence-negative (blue) patients in our internal cohort. **(A)** and **(B)** represent two independently sampled subsets of the same cohort, illustrating consistent clustering patterns across experiments. In both instances, cells from recurred and not recurred patients occupy distinct regions of the embedding space, despite being histologically indistinguishable.

Building on this foundation, PathRosetta achieved substantial performance gains over state-of-the-art AI models and clinical systems. While traditional models treat patches as atomic units, averaging away cellular detail, PathRosetta explicitly preserves cell identity, morphology, and spatial arrangement. This enables the model to learn higher-order phenomena such as tumor–immune interplay, desmoplastic remodeling, and necrosis, which are critical determinants of progression and recurrence. This is quantitatively validated in our results where we showed that the PathRosetta outperformed cell-only and patch-only baselines by large margins (AUC: 0.78 vs. 0.64 and 0.67), demonstrating that recurrence risk is determined not solely by cell presence, but by how these cells interact and are spatially arranged (Supplementary Table S1).

The clinical significance of these improvements is reflected in the model's ability to stratify patients into biologically distinct risk groups. PathRosetta achieved a hazard ratio of 9.54 between predicted high- and low-risk patients, suggesting strong utility for guiding adjuvant therapy and surveillance strategies in early-stage ILA. External validation on independent TCGA and CPTAC LUAD cohorts (AUCs 0.75 and 0.76) further confirms the model's robustness and generalizability across data sources and staining variations.

A key strength of PathRosetta lies in its flexibility that it can construct prognostic "documents" from either the complete cellular ecosystem or from individual cell-type populations. Because it treats each cell as a fundamental unit (analogous to a word in language) and builds hierarchical representations through their spatial relationships, the model allows selective interrogation of which cellular compartments encode the most informative prognostic signals. This capability is intrinsic to our language-inspired architecture, enabling systematic evaluation of the prognostic contribution of each cell type independently while preserving the hierarchical language of pathology at both cellular and spatial scales. Importantly, the cell-type–specific analyses revealed subgroups within different cell types which offer complementary prognostic signals. When we constructed a "document" composed solely of benign epithelial cells and defined its grammar using PathRosetta, this model achieved the highest performance (AUC = 0.75), indicating that non-neoplastic epithelial morphology and organization carry underappreciated prognostic information, potentially reflecting early field effects or adaptive epithelial responses. These results extend the traditional focus on malignant cells alone, reinforcing that cancer progression emerges from the collective dynamics of the tumor ecosystem. To definitively link these computational subgroups to recurrence, we will perform spatial transcriptomics to resolve the specific gene expression programs and molecular differences that drive risk within the pathologically defined cell type.

Interpretability analyses support these biological insights. High-risk predictions consistently localized to regions with solid and micropapillary architecture, nuclear atypia, and necrosis, while low-risk predictions focused on immune-rich or fibrotic areas associated with favorable outcomes. These attention patterns align with established histologic correlates of prognosis, underscoring that PathRosetta's representations are both data-driven and biologically grounded. Representative attention maps for high- and low-risk patients are shown in Supplementary Figures S2 and S3.

Collectively, these findings highlight a conceptual shift in computational pathology, from image-level recognition to language modeling. By explicitly representing tissue as a network of interacting cells, PathRosetta decodes the hierarchical grammar that underlies tumor behavior. This biologically informed design establishes a scalable framework for generalization across cancer types, endpoints, and modalities. Future extensions integrating molecular and spatial transcriptomic data may further connect cellular morphology to gene expression and functional state, deepening our understanding of how tissue organization encodes clinical outcomes.

In summary, PathRosetta reframes histopathology as a decipherable biological language, one in which each cell contributes meaning through its morphology, neighbors, and spatial context. By learning this language, PathRosetta not only advances prognostic modeling but also provides a foundation for interpretable, biologically faithful AI systems in precision oncology.

## Methods

### Study design and patient cohorts

This study utilized a cohort of 189 patients diagnosed with stage I to III ILA, who underwent curative-intent surgical resection at Wake Forest Baptist Comprehensive Cancer Center between 2008 to 2015. All patients were followed up for a minimum of five years. The cohort included 456 H&E-stained WSIs, with a median of two slides per patient. All slides corresponding to a given patient were assigned to the same fold to avoid data leakage. Patient follow-up data included recurrence status and time to recurrence, with 72 patients (38.1%) experiencing recurrence within five years. The study design diagram (Supplementary Fig. S4A) outlines the exclusion criteria and study population, while cohort demographics (Supplementary Fig. S4B) show a balanced distribution of age, with slightly more females (56.6%) than males (40.7%), and the sex of 5 patients (2.6%) unknown. Notably, patients who experienced recurrence had higher rates of lymphovascular invasion (29.2% vs. 7.7%) and higher-grade tumors (84.7% vs. 43.6% for Grade 3 in the new grading system, WHO 5th Ed. [2021]). For external validation, we used two independent public datasets: the TCGA-LUAD (82 cases) and CPTAC-LUAD (18 cases) cohorts. Only cases with available 5-year recurrence data were included.

**PathRosetta Architecture**

PathRosetta implements the "language of histopathology" concept through three core components that mirror natural language processing architectures: (1) multi-scale feature extraction captures both "words" (individual cells at 40×) and "sentences" (tissue patches at 20×), (2) cell-to-patch mapping and fusion creates contextual embeddings by linking cellular "words" with their architectural "sentences," and (3) multi-dimensional attention mechanisms learn complex semantic relationships between tissue regions, analogous to how transformers capture dependencies between sentence elements. This architecture addresses the fundamental limitations of current AI methods by integrating information across multiple scales and explicitly modeling cellular and intercellular characteristics within their biological context. Figure 8 illustrates the overall architecture of PathRosetta.

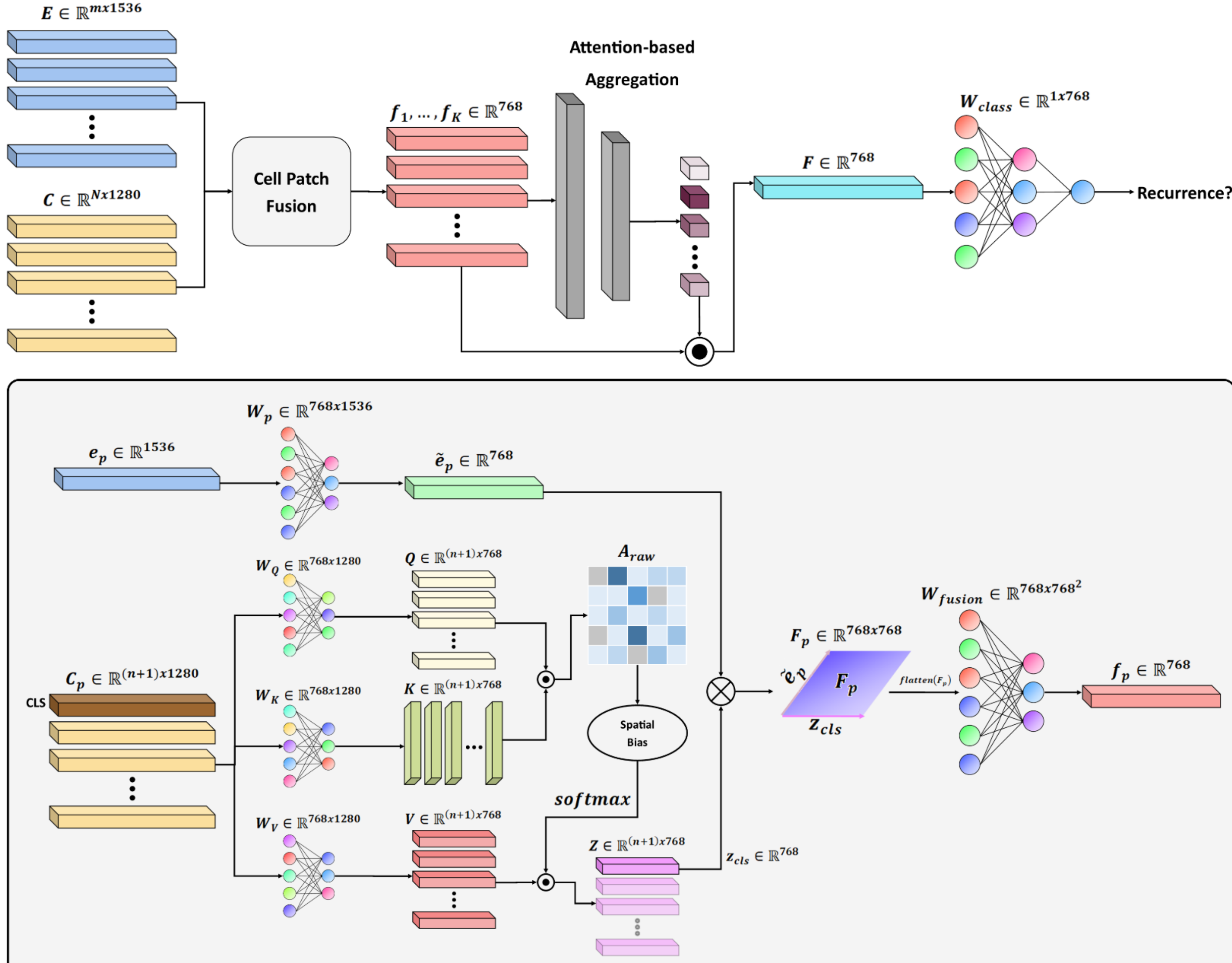

**Figure 8 | Schematic overview of the PathRosetta architecture Top:** PathRosetta integrates patch-level and cell-level representations through a cell–patch fusion mechanism, followed by attention-based aggregation for

whole-slide recurrence prediction. Patch and cell embeddings are projected into a shared latent space and fused to form unified representations which are subsequently aggregated into a slide-level feature vector for recurrence classification. **Bottom:** Detailed architecture of the fusion module. Cell embeddings are combined using a multi-head self-attention mechanism with spatial bias and fused with patch embeddings using outer product. The resulting fused feature map is flattened and passed through a learned fusion layer to generate the final fused representation. This hierarchical design enables PathRosetta to jointly capture cellular morphology, spatial context, and tissue architecture.

**Multi-Scale Feature Extraction** Our multi-scale feature extraction strategy implements the core principle of the pathology language: cells at 40× magnification serve as the fundamental "words" encoding fine-grained morphological and phenotypic information, while tissue patches at 20× magnification represent "sentences" that provide architectural context and spatial organization. This dual-scale approach mirrors how language models require both token-level understanding (individual word meanings) and sequence-level comprehension (sentence structure and context) to achieve semantic understanding. Just as word embeddings capture lexical information while sentence embeddings capture syntactic and semantic relationships, our cellular embeddings encode morphological details while patch embeddings capture tissue architecture and spatial patterns.

We adopted a two-step feature extraction strategy:

- **Patch-Level Feature Extraction (20x):** WSIs were tiled into non-overlapping 256x256 pixel patches at 20x magnification. Using a pre-trained FM, UNI2-h [25] , we extracted a feature vector $e_p \in \mathbb{R}^{1536}$ for each patch $p$, capturing broad morphological and tissue-architectural patterns. We used Trident [37, 38] for tissue segmentation, patch cropping, and feature extraction.
- **Cell-Level Feature Extraction (40x):** Concurrently, cell nuclei were segmented and classified from WSIs at 40× magnification using CellViT++ [35, 36], specifically the SAM-based model [39]. This model yields embeddings $c_i \in \mathbb{R}^{1280}$ for each individual cell $c_i$, encoding fine-grained morphological and phenotypic details across five major cell categories: stromal, inflammatory, neoplastic, dead, and benign epithelial. It also outputs centroid coordinates for each cell. Preprocessing was performed using PathoPatcher [40].

This dual-scale approach (20x for summarizing path level information and 40x for capturing subtle information at cell level) overcomes the limitations of single-scale modeling, enabling the simultaneous utilization of both cellular morphology and tissue architecture.

**Cell-to-Patch Mapping and Spatially Biased Cell Attention** The integration of cellular and tissue-level information implements contextual embedding principles from natural language processing, where the meaning of individual elements depends critically on their surrounding context. In PathRosetta, a cell's biological significance, like a word's semantic meaning, is determined not only by its intrinsic morphological features but also by its spatial neighbors and architectural context. The spatially biased attention mechanism we introduce mimics how attention mechanisms in language models allow words to attend to relevant context, enabling our framework to automatically learn which cellular neighborhoods are most informative for recurrence prediction.

**Cell-to-Patch Mapping** We established a spatial correspondence between cells and patches by mapping each cell to the patch containing its centroid. Formally, for each patch $p$, we define the set of corresponding cells $as$:

$$C_p = \{c_i \in C | centroid(c_i) \in p\},$$

where $C_p$ is the set of cells whose centroids lie within the spatial bounds of patch $p$, and $C$ is the full set of cells. This mapping preserves spatial context, linking cells to the patch that contains them.

**Spatially Biased Cell Self-Attention** Each patch embedding $e_p \in \mathbb{R}^{1536}$, was linearly projected to $\mathbb{R}^{768}$ via a learned projection:

$$\tilde{e}_p = W_p e_p,\ where\ W_p \in \mathbb{R}^{768\times1536}$$

Each patch $p$ contains $c$ cells, we denote the cell embeddings as $c_1, c_2, c_3, \ldots, c_c \in \mathbb{R}^{1280}$. Each $c_i$ is passed through LayerNorm both before and after attention. We aggregate these variable-length cell embeddings using a self-attention mechanism with a learnable CLS token $e_{CLS} \in \mathbb{R}^{1280}$. The input sequence is:

$$C_p = [e_{CLS}, c_1, \ldots, c_c] \in \mathbb{R}^{(n+1)\times 1280}$$

We apply learned projections:

$$Q = W_Q \cdot X, \qquad K = W_K \cdot X, \qquad V = W_V \cdot X,$$

where $W_Q, W_K, W_V \in \mathbb{R}^{768\times 1280}$.

We computed the raw attention scores between all elements:

$$A_{raw} = \frac{(Q \cdot K^T)}{\sqrt{d_K}},$$

where $d_K$ is the dimension of the key vectors i.e., 768. Then, we incorporated spatial information by subtracting the physical distance between cell centroids from the attention scores:

$$A_{spatial(i,j)} = A_{raw(i,j)} - dist\left(centroid_i, centroid_j\right),$$

where $dist(\cdot,\cdot)$ is the Euclidean distance between cell centroids. This spatial bias implements the grammatical rules of pathology language, where cellular proximity defines semantic relationships—just as word order and adjacency create meaning in natural language, spatial arrangements of cells encode biological function and prognostic significance. Cells in close proximity form functional neighborhoods that act as coherent semantic units within the tissue microenvironment. This enables the framework to capture local cellular neighborhoods and their collective behavior, which is crucial for understanding TME characteristics and their role in cancer progression. We then applied softmax normalization to obtain the final attention weights:

$$A = softmax(A_{spatial})$$

The output of the self-attention layer is computed by:

$$Z = A \cdot V$$

The updated CLS token $z_{CLS} \in \mathbb{R}^{768}$ is extracted from the first row of $Z$, which now contains aggregated information from all cells within the patch, with greater influence from spatially proximal cells.

This spatially biased self-attention mechanism allows the framework to selectively focus on informative cells while capturing cell-cell interactions within each patch, addressing the homogeneous treatment of cells in current approaches.

**Cross-Scale Fusion** Cross-scale fusion represents the critical step where "word-level" cellular information is integrated with "sentence-level" tissue architecture, analogous to how transformer models combine token embeddings with positional and contextual information. This fusion captures the biological reality that cellular behavior cannot be understood in isolation—just as word meaning depends on sentence context, cellular significance emerges from the interplay between intrinsic cell properties and their tissue microenvironment. The outer product operation we employ creates a rich representation space that captures all possible interactions between cellular morphology and architectural context.

For each patch $p$, we have a patch embedding vector $\tilde{e}_p \in \mathbb{R}^{768}$ and an aggregated cell embedding vector $z_{CLS} \in \mathbb{R}^{768}$, we computed the outer product between these vectors:

$$F_p = \tilde{e}_p \otimes z_{CLS} \in \mathbb{R}^{768\times 768}$$

This matrix captures all pairwise multiplicative interactions between patch and cell features. To make this representation computationally tractable, we flattened the matrix and projected it back to 768 dimensions using a learnable projection matrix $W_{fusion}$:

$$f_p = W_{fusion} \cdot flatten(F_p), \quad W_{fusion} \in \mathbb{R}^{768 \times 768^2}$$

where $f_p \in \mathbb{R}^{768}$ is the final fused representation for patch $p$.

This fusion mechanism enables rich cross-scale interactions that would be missed by simpler concatenation or addition operations, allowing the framework to capture how cellular characteristics interact with tissue architecture to influence recurrence risk.

**Multi-Dimensional Attention MIL** The fused embeddings $f_1, \dots, f_K \in \mathbb{R}^{768}$ from a WSI form a "bag" of $K$ instances. We employed AttMIL to aggregate these instance features into a slide-level representation for predicting recurrence. This concept is visualized in Figure 2(B).

**Conventional 1D Attention (AttMIL)** As a baseline and part of our ablation, we used the standard AttMIL, where instance embeddings $h_k$ (our $f_k$) are projected onto a 1D space, see the equation below for reference, to get attention weights $\alpha_k$, which are then used to compute a weighted sum of instance embeddings.

$$\alpha_k = \frac{\exp\{w^{\mathrm{T}}\left(\tanh\left(V{h_k}^T\right) \odot sigmoid\left(U{h_k}^T\right)\right)\}}{\sum_{j=1}^{K} \exp\{w^{\mathrm{T}}\left(\tanh\left(V{h_j}^T\right) \odot sigmoid\left(U{h_j}^T\right)\right)\}}$$

where $V, U$ are learnable subspaces, and $\odot$ denotes element-wise multiplication. The scalar $w^{\mathrm{T}}(\cdot)$ collapses the transformed instances into a single attention score. The reliance on a single weight vector $w$ can produce identical attention scores for instances with divergent feature interactions when projected onto the 1D subspace defined by $w$.

**Novel 2D and 3D Attention Projections** To overcome the limitations of 1D projection, we extended this by projecting $h_k$ onto a 2D or 3D plane/space using two $(w_1, w_2)$ or three $(w_1, w_2, w_3)$ learnable weight vectors, respectively. This can be written mathematically as:

$$\alpha_k = \frac{\exp\{\left\|W^{\mathrm{T}}\left(\tanh\left(V{h_k}^T\right) \odot sigmoid\left(U{h_k}^T\right)\right)\right\|_F\}}{\sum_{j=1}^{K} \exp\{\left\|W^{\mathrm{T}}\left(\tanh\left(V{h_j}^T\right) \odot sigmoid\left(U{h_j}^T\right)\right)\right\|_F\}}$$

where $\|\cdot\|_F$ is the Frobenius norm. This allows the framework to assign importance based on a richer, multi-faceted representation of each instance.

**Cell-Type Specific Models** Cell-type-specific modeling reflects the concept that different cellular populations represent distinct "vocabularies" within the pathology language, each with specialized semantic roles in the TME. Just as domain-specific language models excel in specialized contexts (medical texts, legal documents, scientific literature), our cell-type-specific models become experts in interpreting the unique contributions of different cellular populations to recurrence risk. The ensemble approach combines these specialized "linguistic perspectives," recognizing that comprehensive understanding of tissue biology—like natural language comprehension—benefits from integrating multiple specialized knowledge domains.

Let the five major cell types be denoted by the set:

$$T = \{stromal, inflammatory, neoplastic, dead, benign\ epithelial\}$$

**Cell-Type Specific Subsets** For each cell type $t \in T$, we constructed a subset of each WSI as follows:

Let $C_t \subseteq C$ be the set of all cells of type $t$ present in a WSI.

Using the spatial mapping dictionary, we selected the set of patches $P_t \subseteq P$, where each patch $p \in P_t$ satisfies:

$$C_p \cap C_t \neq \emptyset,$$

meaning the patch contains at least one cell of type $t$.

Each model $M_t$ was then trained using the subset $S_t = \{P_t, C_t\}$, resulting in five specialized models focused on learning the recurrence-related patterns associated with a specific cell type.

A sixth model, denoted $M_{all}$, was trained using all cell embeddings and all patches that contained at least one cell of any type:

$$P_{all} = \{p \in P | C_p \neq \emptyset\}$$

This model captures holistic cellular heterogeneity and serves as a comprehensive multi-cell-type baseline.

**Ensemble Integration via Patient-Level Majority Voting** To integrate predictions across cell-type-specific perspectives, we employed a patient-level ensemble approach based on majority voting.

Let a patient $i$ have $S_i$ WSIs, and let $M = \{M_1, M_2, \ldots, M_6\}$ be the six trained models (five cell-type-specific and one multi-type model). Each model $M_m \in M$ outputs a probability score ${p_{i,s}}^{(m)} \in [0, 1]$ for slide $s \in \{1, \ldots, S_i\}$ of patient $i$.

The aggregated patient-level probability for model $M_m$ is computed as the average over all WSIs:

$$\tilde{p}_i{}^{(m)} = \frac{1}{S_i} \sum_{s=1}^{S_i} {p_{i,s}}^{(m)}$$

The final patient-level prediction is then obtained by averaging the probabilities across all models:

$$\tilde{p}_i = \frac{1}{|M_i|} \sum_{m \in M_i} \tilde{p}_i{}^{(m)},$$

where $M_i \subseteq M$ includes only the models applicable to patient $i$ (i.e., for which at least one relevant cell is found).

The final binary decision is made using a threshold $\tau \in [0,1]$ ($\tau = 0.5$):

$$\hat{y}_i = \begin{cases} 1, & if\ \tilde{p}_i \geq \tau \\ 0, & otherwise \end{cases}$$

This ensemble strategy leverages the distinct predictive power of each cell type while combining them to improve generalizability and robustness. It also reflects the biological heterogeneity of the TME, where different cell populations contribute uniquely to tumor progression and recurrence.

**Training and optimization**

Models were trained using five-fold cross-validation on NVIDIA A100 GPUs for up to 100 epochs with early stopping based on validation performance. Optimization was performed using the Adam optimizer with an initial learning rate of $5 \times 10^{-5}$, weight decay of $1 \times 10^{-5}$, and binary cross-entropy loss with L2 regularization. To ensure clinically balanced predictions, we introduced a harmonized stopping criterion that maximizes the harmonic mean of sensitivity and specificity on the validation set. The model corresponding to the best validation epoch was used for testing.

**Baseline and comparative methods**

We compared PathRosetta against several state-of-the-art computational pathology methods, including AttMIL [41], CLAM [42], TransMIL [43], CAMIL [44], DAMIL [20], and Kim et al. (EfficientNet-b0 + AttMIL) [45]. Clinical baselines included AJCC stage, dominant pattern grading, and the IASLC grading system. For clinical systems, AUCs were computed using machine learning models like logistic regression, random forest, xgboost, support vector machine, etc. and C-indices from Cox proportional hazards models.

**Evaluation metrics**

Performance was evaluated using AUC, accuracy, sensitivity, specificity, and C-index. Statistical significance was assessed using the log-rank test for Kaplan–Meier analyses and 95% confidence intervals for hazard ratios (Cox models).

All metrics represent mean ± standard deviation across test folds. External validation results on TCGA and CPTAC cohorts were computed on the model having the best performance on validation set and are reported separately (Supplementary Table S6).

**Statistical analysis**

All statistical analyses were performed using Python 3.10 with scikit-learn 1.3, lifelines 0.27, and statsmodels 0.14. A p-value $< 0.05$ was considered statistically significant.

**Author contributions**

A. R. A. performed data preprocessing, designed and conducted experiments, validation, and manuscript writing. U. S. and Z. S. provided feedback on the study design and validation and also edited the manuscript. W. L, F. X., and J. R. performed data collection, provided clinical assessment, and contributed to the manuscript editing. W. C. and M.K.K.N. conceptualized and designed the study, provided study validation, supervised the research, and contributed to the manuscript writing and editing.

**Acknowledgments**

The authors gratefully acknowledge the Ohio Supercomputer Center for providing high-performance computing resources as part of its contract with The Ohio State University College of Medicine. We also thank the Department of Pathology and the Comprehensive Cancer Center at The Ohio State University for their support.

**Funding**

The project described was supported in part by R01 CA276301 (PIs: Niazi and Chen) from the National Cancer Institute, R01 DC020715 (PIs: Moberly and Gurcan) from the National Institute on Deafness and Other Communication Disorders, and R01 HL177046-01 (PIs: Hachem, Gurcan, and Michelson) from the National Heart, Lung, and Blood Institute. The project was also supported partly by Pelotonia under IRP CC13702 (PIs: Niazi, Vilgelm, and Roy), The Ohio State University Department of Pathology and Comprehensive Cancer Center. The content is solely the responsibility of the authors and does not necessarily represent the official views of the National Cancer Institute or National Institutes of Health or The Ohio State University.

**Data Availability**

The data that support the findings of this study are available from Wake Forest Baptist Comprehensive Cancer Center, but restrictions apply to the availability of these data, which were used under license for the current study, and so are not publicly available. Data are, however, available from the authors upon reasonable request and with permission of Wake Forest Baptist Comprehensive Cancer Center. TCGA-LUAD is available through the National Cancer Institute's Genomic Data Commons (GDC) portal (https://portal.gdc.cancer.gov) and CPTAC-LUAD is available through the Cancer Imaging Archive (TCIA) portal (https://www.cancerimagingarchive.net/collection/cptac-luad/).

### Code Availability

The underlying code for this study is available in AI4Path PathRosetta repository and can be accessed via this link https://github.com/AI4Path-Lab/PathRosetta.

### Ethics Approval and Consent to Participate

This study was reviewed and approved by the Institutional Review Board of Wake Forest University Health Sciences (IRB #00074626). All procedures involving human data were conducted in accordance with the ethical standards of the institutional research committee, the national research regulations, and with the 1964 Declaration of Helsinki and its later amendments. Given the retrospective nature of the study and because data were archival and de-identified, the requirement for informed consent to participate was formally waived by the Institutional Review Board.

## Supplementary Materials

### S1 Ablation Study

To rigorously evaluate the contributions of different components in our framework, we conducted a comprehensive ablation study across various configurations: 1D Projection (baseline), 2D Projection, 3D Projection, 1D without patch embeddings, and 2D without patch embeddings. These configurations test different aspects of our architecture: 1D, 2D, and 3D Projection variants modify the dimensionality of the attention mechanism (projecting instance features onto 1-dimensional, 2-dimensional, or 3-dimensional spaces for attention weight calculation), while the "without patch embeddings" variants remove tissue-level context by using only cell-level information without the corresponding patch-level features from the same spatial regions. Performance was assessed using accuracy, sensitivity, specificity, and AUC on both validation and test sets.

### S1.1 Validation Performance Comparison

The 1D Projection (baseline) achieved a mean validation AUC of 67.63 ± 3.34, with sensitivity and specificity at 59.92 ± 2.91 and 63.92 ± 4.74, respectively. The 2D Projection demonstrated significant improvements, achieving a 3.6% higher mean validation AUC (70.06 ± 1.95) than the baseline. Notably, the 2D variant improved sensitivity by 12.7% (67.55 ± 7.37 vs. 59.92 ± 2.91), confirming its ability to better model feature interactions through orthogonal attention mechanisms. Expanding to 3D Projection did not yield further gains; instead, validation AUC dropped by 3.3% (67.61 ± 6.93) compared to 2D, with sensitivity and specificity variance spiking (> ±10.01 std), indicating overparameterization and training instability.

The most severe degradation occurred when patch embeddings were removed. The 1D variant without patches collapsed to a 15.3% lower validation AUC (57.31 ± 3.83) than the baseline, while the 2D variant preserved marginally better performance (62.08 ± 4.83 AUC), still 8.4% worse than standard 2D. This underscores that patch embeddings contribute critically to model performance, meaning that they capture some complementary information to the cell-level representations.

### S1.2 Test Set Confirmation

Test results mirrored validation trends: 2D Projection achieved the highest mean AUC (74.28 ± 3.07), outperforming 1D by 4.9% (70.84 ± 2.48) and 3D by 3.9% (71.50 ± 2.00). Specificity stability also improved (2D std: ±5.78 vs. 1D: ±6.38), reinforcing its robustness. Removing patch embeddings consistently degraded test metrics, validating our design choice to retain them.

The final model (2D attention with patch embeddings) was selected based on validation metrics, with test results and cell-specific analyses further corroborating its advantage, see Table S1 for detailed results. This systematic evaluation ensures that our design decisions are empirically grounded, maximizing both discriminative power and robustness.

***Table S1:*** *Performance comparison across ablation settings, showing validation and test results (mean ± std) for models with different projection dimensions (1D, 2D, 3D) and patch embedding configurations. Bold values highlight top-performing configurations per metric, while the underlined show the second best. P.E. denotes patch embeddings, A.P. denotes attention projection, Acc stands for accuracy, Sen for sensitivity, and Spe for specificity.*

| **Ablation Settings** | | **Validation Results** | | | | **Test Results** | | | |
|---|---|---|---|---|---|---|---|---|---|
| **P.E.** | **A.P.** | **Accuracy** | **Sensitivity** | **Specificity** | **AUC** | **Accuracy** | **Sensitivity** | **Specificity** | **AUC** |
| ✓ | 1D | 62.38±4.03 | 59.91±2.91 | 63.90±4.74 | 67.63±3.34 | 65.46±3.21 | **67.30±5.51** | 63.90±6.38 | 70.83±2.48 |
| ✓ | 2D | **65.74±3.22** | 67.55±7.37 | **64.34±6.94** | **70.06±1.95** | **68.60±1.62** | 61.89±6.17 | **73.32±5.78** | **74.28±3.07** |
| ✓ | 3D | 62.08±8.04 | **67.68±10.01** | 58.52±10.25 | 67.61±6.93 | 64.16±2.30 | 66.94±8.37 | 62.25±9.16 | 71.50±2.00 |
| ✖ | 1D | 57.60±5.54 | 54.60±7.64 | 59.37±5.25 | 57.30±3.83 | 61.18±0.85 | 55.75±3.56 | 64.89±3.67 | 63.52±2.26 |
| ✖ | 2D | 58.53±5.38 | 65.15±12.7 | 54.07±7.29 | 62.07±4.83 | 61.58±3.61 | 64.34±5.59 | 59.90±3.93 | 64.28±5.27 |

***Table S2:*** *Performance comparison of state-of-the-art MIL methods with our best performing model. Metrics show mean ± standard deviation across test folds. *DAMIL uses a different encoder (CTransPath), and data folds compared to other methods. Best performance in each metric is bolded, while second-best is underlined.*

| Model | Accuracy (%) | Sensitivity (%) | Specificity (%) | AUC (%) |
|---|---|---|---|---|
| **AttMIL** | 61.55 ± 8.23 | 55.77 ± 12.04 | 65.39 ± 7.79 | 66.57 ± 5.55 |
| **TransMIL** | 58.85 ± 5.14 | 54.75 ± 5.29 | 61.26 ± 9.2 | 62.29 ± 3.85 |
| **CLAM** | 61.13 ± 7.52 | 56.52 ± 12.89 | 64.56 ± 12.18 | 66.13 ± 5.90 |
| **CAMIL** | 66.77 ± 3.75 | 57.50 ± 7.17 | 72.60 ± 3.63 | 67.47 ± 5.70 |
| **Kim et al.** | 61.84 ± 7.18 | **68.00 ± 11.6** | 58.16 ± 17.93 | 66.81 ± 5.54 |
| **DAMIL*** | 63.50 ± 3.10 | 53.00 ± 7.90 | 69.30 ± 6.60 | 64.90 ± 1.20 |
| **PathRosetta (ours)** | **71.72 ± 4.82** | 65.81 ± 7.39 | **75.47 ± 5.38** | **77.84 ± 3.99** |

***Table S3:*** *Performance comparison of cell-type-specific models, the All-Cells model, and the Combined Model for ILA recurrence prediction. Metrics include accuracy, sensitivity, specificity, and AUC (mean ± standard deviation across test folds). The Combined Model (ensemble) achieves the highest performance across all metrics (bold). The second-best results are underlined.*

| Model | Accuracy (%) | Sensitivity (%) | Specificity (%) | AUC (%) |
|---|---|---|---|---|
| **Stromal Cells** | 63.81 ± 4.40 | 62.89 ± 7.27 | 64.26 ± 7.63 | 71.69 ± 6.31 |
| **Inflammatory Cells** | 64.12 ± 3.50 | 62.09 ± 5.50 | 65.27 ± 3.84 | 70.27 ± 6.49 |
| **Neoplastic Cells** | 65.64 ± 4.78 | 63.36 ± 4.69 | 67.13 ± 5.52 | 71.92 ± 3.26 |
| **Dead Cells** | 67.30 ± 1.73 | 64.38 ± 10.63 | 69.14 ± 7.80 | 72.99 ± 3.89 |
| **Benign Epithelial Cells** | 69.14 ± 2.84 | 64.94 ± 4.39 | 71.24 ± 4.99 | 74.98 ± 3.36 |
| **All-Cells** | 68.60 ± 1.62 | 61.89 ± 6.17 | 73.32 ± 5.78 | 74.28 ± 3.07 |
| **Combined Model** | **71.72 ± 4.82** | **65.81 ± 7.39** | **75.47 ± 5.38** | **77.84 ± 3.99** |

***Table S4:*** *HR and C-Index of all models including cell-type specific, all-cell and combined models. The table also shows the 95% confidence interval of the HRs with a p-value < 0.005, as well as the standard deviation of C-Index across folds.*

| Model | HR | C-Index |
|---|---|---|
| **Stromal Cell** | 5.44 [2.69–11.02] | 0.671 ± 0.05 |
| **Inflammatory Cell** | 5.06 [2.41–10.62] | 0.666 ± 0.02 |
| **Neoplastic Cell** | 4.98 [2.46–10.06] | 0.663 ± 0.04 |
| **Dead Cell** | 5.69 [2.96–10.96] | 0.688 ± 0.02 |
| **Benign Epithelial Cell** | 6.04 [3.15–11.56] | **0.708 ± 0.02** |
| **All-Cells** | 7.70 [3.80–15.60] | 0.698 ± 0.03 |
| **Combined Model** | **9.54 [4.34–20.98]** | 0.704 ± 0.02 |

***Table S5:*** *Subgroup-wise performance metrics of the model across demographic and clinical variables. N denotes the number of patients in each subgroup. The results are reported by combining the test splits from all cross-validation folds, ensuring complete coverage of the dataset. "Old Grade" refers to the traditional predominant pattern–based grading system in WHO 4$^{th}$ Ed. (2015), whereas "New Grade" corresponds to the IASLC grading system in WHO 5$^{th}$ Ed. (2021). Metrics reported include AUC, accuracy, sensitivity, and specificity, computed separately for each subgroup to evaluate model fairness and robustness.*

| Subgroup | N | Accuracy | Sensitivity | Specificity | AUC |
|---|---|---|---|---|---|
| **White** | 149 | 71.14 | 66.67 | 73.91 | 75.93 |
| **African American** | 20 | 65.00 | 75.00 | 58.33 | 77.08 |
| **Age < 65** | 100 | 66.00 | 64.71 | 66.67 | 69.56 |
| **Age >= 65** | 89 | 74.16 | 65.79 | 80.39 | 80.70 |
| **Female** | 107 | 69.16 | 61.29 | 72.37 | 71.43 |
| **Male** | 77 | 70.13 | 69.23 | 71.05 | 75.44 |
| **New Grade G1** | 24 | 87.50 | 100.00 | 86.96 | 86.96 |
| **New Grade G2** | 53 | 73.58 | 70.00 | 74.42 | 73.26 |
| **New Grade G3** | 112 | 64.29 | 63.93 | 64.71 | 69.98 |
| **Old Grade G1** | 57 | 75.44 | 60.00 | 78.72 | 69.36 |
| **Old Grade G2** | 102 | 70.59 | 70.83 | 70.37 | 77.78 |
| **Old Grade G3** | 28 | 60.71 | 53.85 | 66.67 | 62.05 |
| **Stage I** | 130 | 73.85 | 64.86 | 77.42 | 75.12 |
| **Stage II** | 41 | 65.85 | 69.57 | 61.11 | 69.57 |
| **Stage III** | 15 | 53.33 | 58.33 | 33.33 | 52.78 |

***Table S6:*** *Performance comparison of the best-performing models, selected based on validation results from five-fold cross-validation on the WFBCCC cohort, on the TCGA-LUAD and CPTAC-LUAD external test sets. The table reports key performance metrics, including AUC, accuracy, sensitivity and specificity, comparing state-of-the-art MIL methods and the proposed PathRosetta framework.*

| | **TCGA-LUAD** | | | | **TCGA-CPTAC** | | | |
|---|---|---|---|---|---|---|---|---|
| **Model** | **Accuracy (%)** | **Sensitivity (%)** | **Specificity (%)** | **AUC (%)** | **Accuracy (%)** | **Sensitivity (%)** | **Specificity (%)** | **AUC (%)** |
| **AttMIL** | 69.51 | **90.38** | 33.33 | 65.90 | 50.00 | 44.44 | 55.56 | 45.68 |
| **TransMIL** | 59.76 | 61.54 | 56.67 | 63.59 | 55.56 | 11.11 | **100.0** | 62.96 |
| **CLAM** | 68.29 | 86.54 | 36.67 | 65.64 | 55.56 | **66.67** | 44.44 | 62.96 |
| **CAMIL** | 50.00 | 26.92 | **90.00** | 63.46 | 33.33 | 00.00 | 66.67 | 38.27 |
| **Kim et al.** | 68.29 | 82.69 | 43.33 | 69.68 | 55.56 | 22.22 | 88.89 | 60.49 |
| **PathRosetta (ours)** | **74.39** | 75.00 | 73.33 | **75.96** | **72.22** | **66.67** | 77.78 | **76.54** |

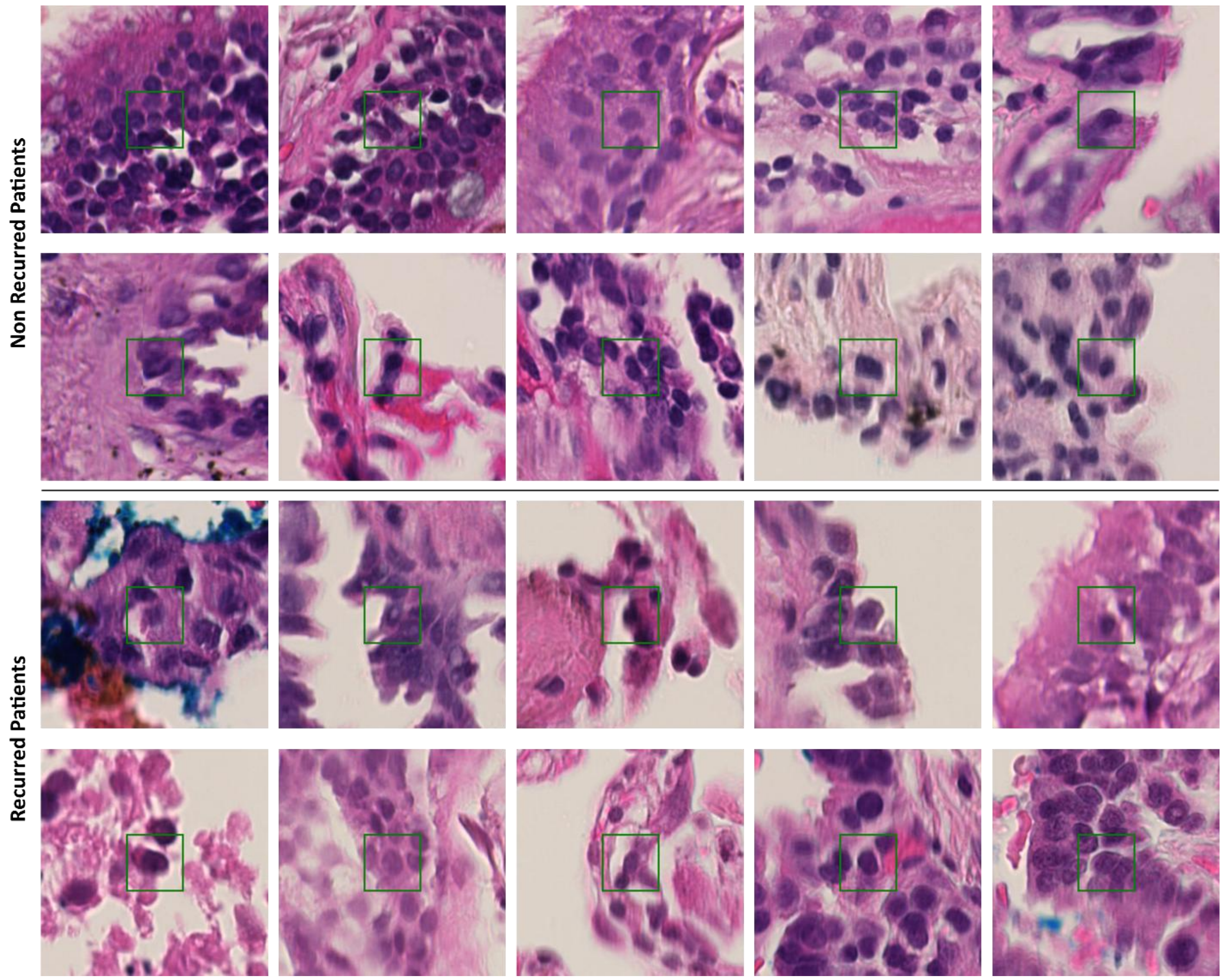

**Figure S1:** This figure displays sample cells from recurrent and non-recurrent patients, encoded differently by CellViT++, leading to their distinct spatial separation in the 3D t-SNE embedding space (Fig. 7).

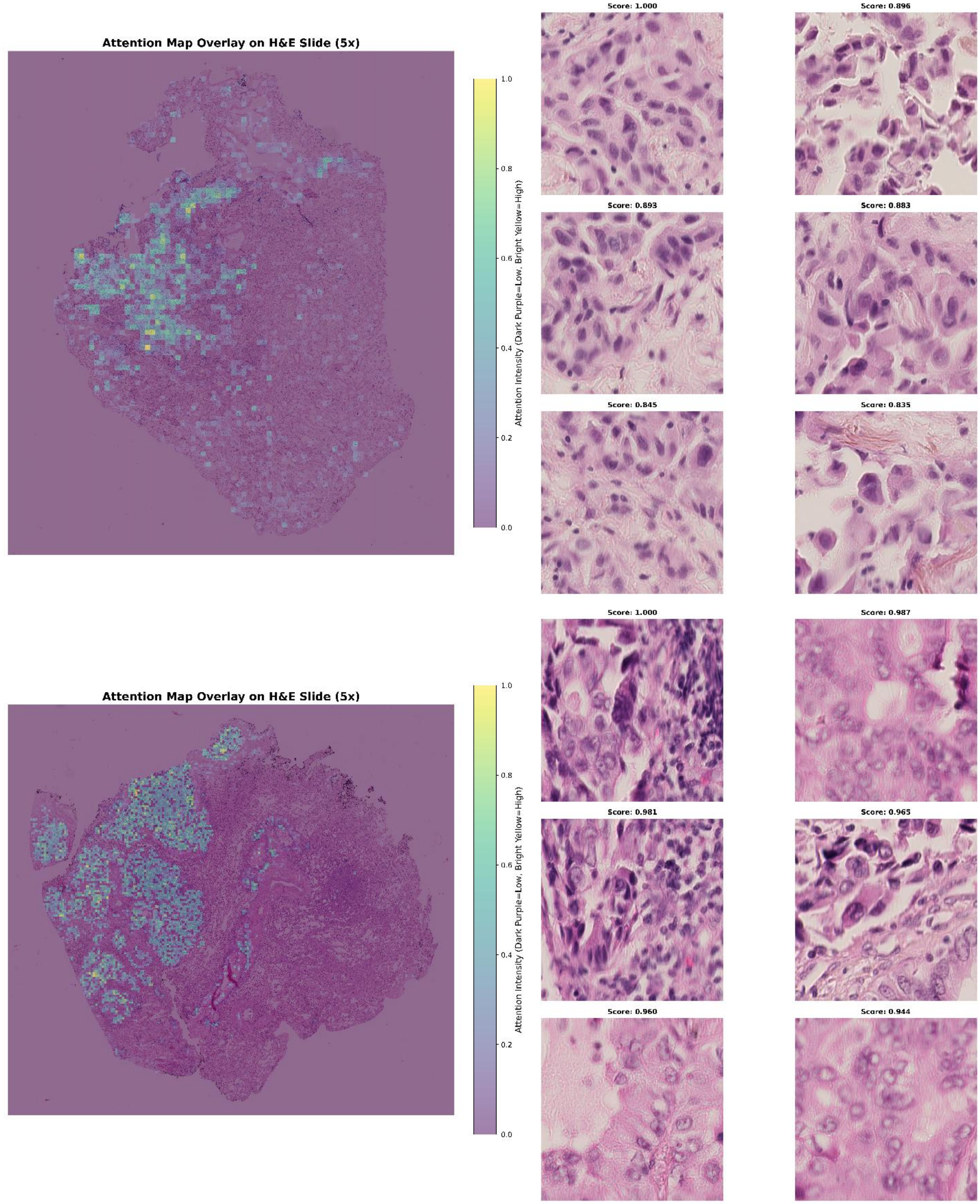

**Figure S2:** Attention maps for correctly predicted recurred patients. In each row, a representative WSI's attention map is shown alongside its topmost highly attended patches. These patterns are all consistent with known poor prognostic indicators.

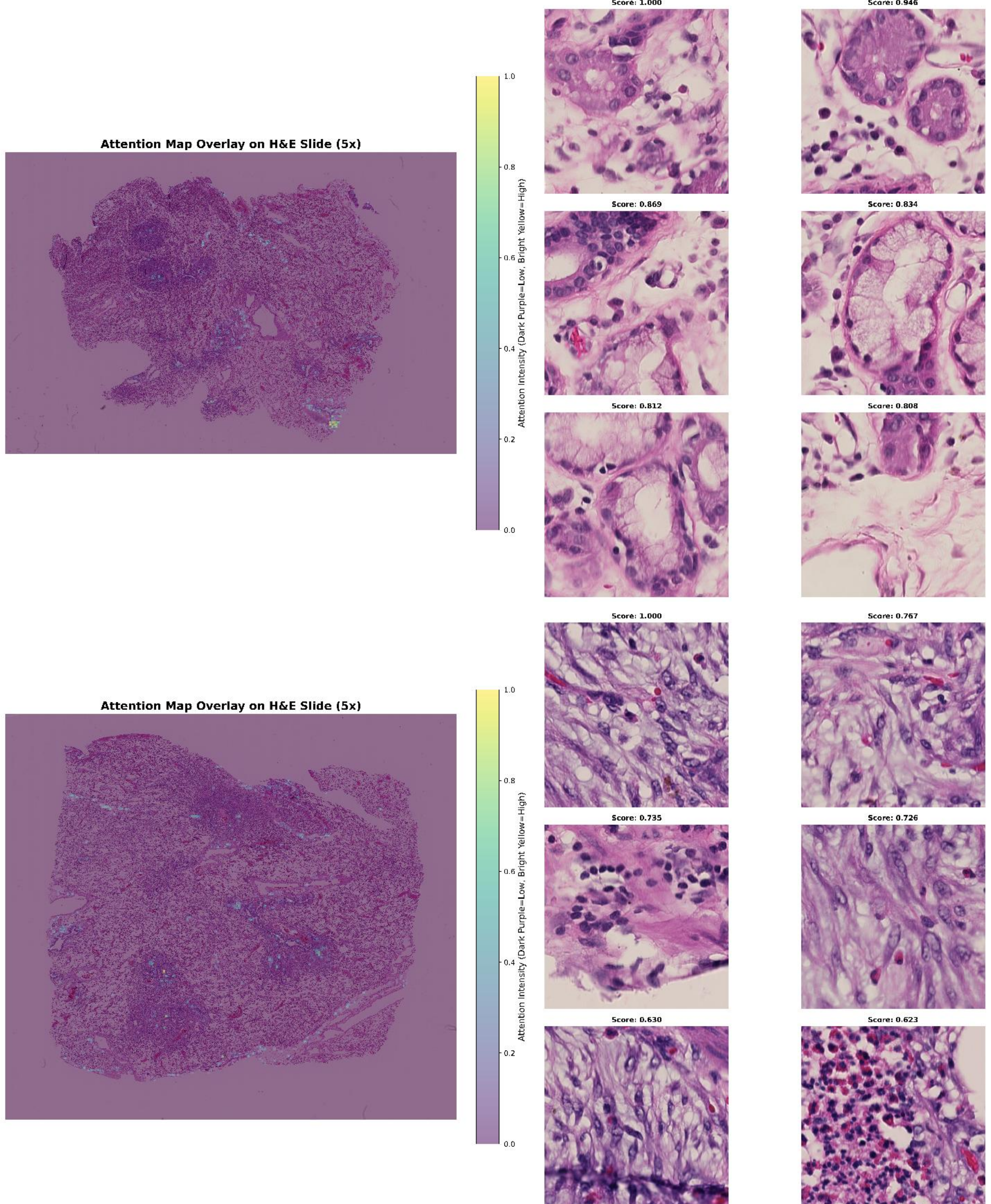

**Figure S3:** Attention maps for correctly predicted non-recurred patients. These patterns suggest the model integrates microenvironmental cues in its risk assessment.

**A**

**Inclusion Criteria:**
- >18 years old
- Histologically confirmed Stage I-III ILA
- Underwent surgical resection at WFBCCC between 2008 to 2015
- 5-Year follow-up data available

**Exclusion Criteria:**
- With previous lung cancer
- With other cancers or serious diseases
- Mucinous adenocarcinoma
- Lost follow-up in 5 years

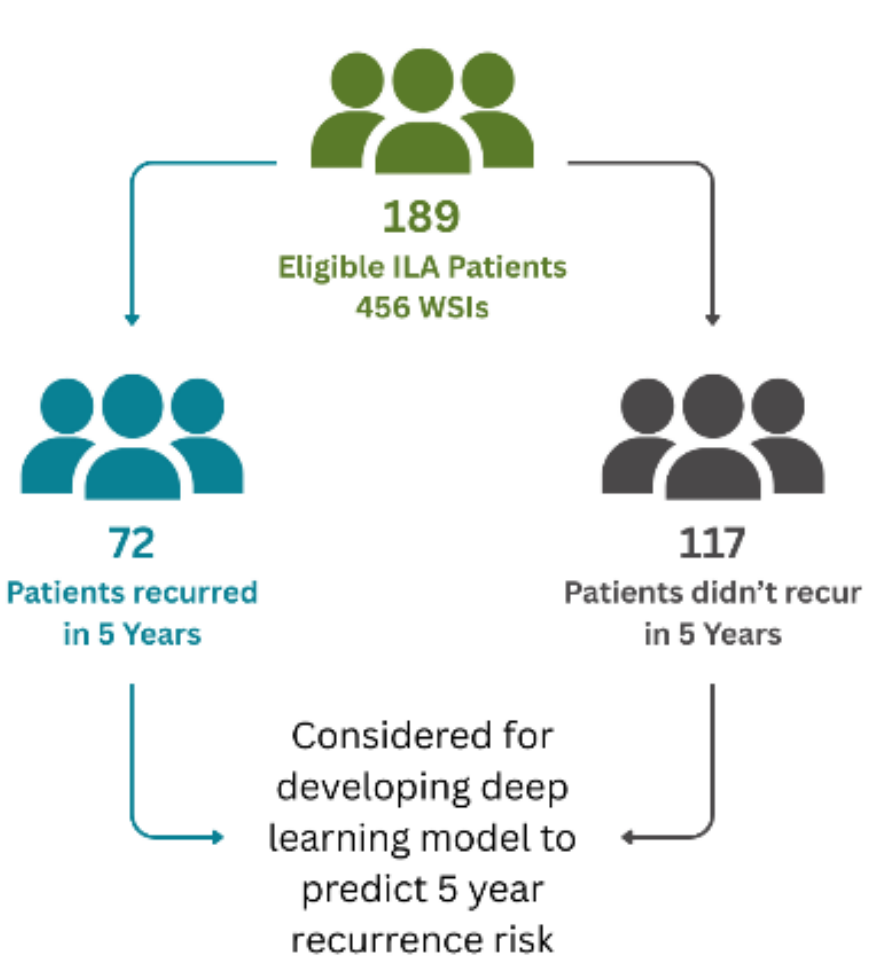

**B**

| Characteristic | All (n=189) | No recurrence (n=117) | Recurrence (n=72) |
|---|---|---|---|
| **Age at diagnosis** | | | |
| Median (min, max) | 63.3 (39.0, 88.0) | 63.0 (43.0, 88.0) | 65.5 (39.0, 84.0) |
| **Race** | | | |
| African American | 20 (10.6%) | 12 (10.3%) | 8 (11.1%) |
| American Indian or Alaska Native | 6 (3.2%) | 3 (2.6%) | 3 (4.2%) |
| Asian | 1 (0.5%) | 1 (0.9%) | 0 (0.0%) |
| Black | 3 (1.6%) | 3 (2.6%) | 0 (0.0%) |
| Hispanic | 3 (1.6%) | 2 (1.7%) | 1 (1.4%) |
| Native Hawaiian or Other Pacific Islander | 1 (0.5%) | 1 (0.9%) | 0 (0.0%) |
| Other | 6 (3.2%) | 3 (2.6%) | 3 (4.2%) |
| White | 149 (78.8%) | 92 (78.6%) | 57 (79.2%) |
| **Sex** | | | |
| Female | 107 (56.6%) | 76 (65.0%) | 31 (43.1%) |
| Male | 77 (40.7%) | 38 (32.5%) | 39 (54.2%) |
| **Death** | | | |
| Dead | 44 (23.3%) | 1 (0.9%) | 43 (59.7%) |
| Alive | 145 (76.7%) | 116 (99.1%) | 29 (40.3%) |
| **Survival (month)** | | | |
| Median (min, max) | 60.0 (4.0, 60.0) | 60.0 (25.0, 60.0) | 47.0 (4.0, 60.0) |
| **AJCC Stage** | | | |
| Median (min, max) | 2.0 (1.0, 7.0) | 1.0 (1.0, 7.0) | 2.0 (1.0, 6.0) |
| **Predominant Pattern-Based Grade** | | | |
| Grade 1 | 57 (30.2%) | 47 (40.2%) | 10 (13.9%) |
| Grade 2 | 102 (54.0%) | 54 (46.2%) | 48 (66.7%) |
| Grade 3 | 30 (15.9%) | 16 (13.7%) | 14 (19.4%) |
| **IASLC Grade** | | | |
| Grade 1 | 24 (12.7%) | 23 (19.7%) | 1 (1.4%) |
| Grade 2 | 53 (28.0%) | 43 (36.8%) | 10 (13.9%) |
| Grade 3 | 112 (59.3%) | 51 (43.6%) | 61 (84.7%) |
| **Intratumoral lymphocytes** | | | |
| Present | 92 (48.7%) | 58 (49.6%) | 34 (47.2%) |
| Absent | 97 (51.3%) | 59 (50.4%) | 38 (52.8%) |
| **Lymphovascular invasion** | | | |
| Absent | 159 (84.1%) | 108 (92.3%) | 51 (70.8%) |
| Present | 30 (15.9%) | 9 (7.7%) | 21 (29.2%) |

**Figure S4: (A)** Study population characteristics and cohort design. This shows patient selection criteria for invasive lung adenocarcinoma cases from 2008-2015, including exclusion criteria (previous lung cancer, other cancers or serious diseases (life-threatening conditions), mucinous adenocarcinoma, and lost follow-up within 5 years). The final study population comprised 189 patients with 456 whole slide images, stratified into recurrence (patients = 72, WSIs = 170) and non-recurrence (patients = 117, WSIs = 286) groups based on 5-year follow-up data. **(B)** Demographic and clinical characteristics of patients with invasive lung adenocarcinoma examined in this study. Data are presented for the entire cohort (n=189) and stratified by recurrence status (no recurrence, n=117; recurrence, n=72). Variables include age at diagnosis, race, sex, survival outcomes, AJCC stage, histological grade (old and new classification systems), intratumoral lymphocytes, and lymphovascular invasion status.